\documentclass{article} 

\usepackage[T1]{fontenc}
\usepackage{tgtermes}

\usepackage{scalefnt,letltxmacro}
\LetLtxMacro{\oldtextsc}{\textsc}
\renewcommand{\textsc}[1]{\oldtextsc{\scalefont{1.10}#1}}

\usepackage{amsmath, amsthm, amsfonts}
\usepackage{mathtools}

\usepackage[
  paper  = letterpaper,
  left   = 1.50in,
  right  = 1.50in,
  top    = 1.0in,
  bottom = 1.0in,
  ]{geometry}
\usepackage[english]{babel}
\usepackage[parfill]{parskip}
\usepackage{afterpage}
\usepackage{enumitem}
\usepackage{framed}
\usepackage{xspace}

\usepackage{lineno}

\usepackage{ragged2e}

\newcounter{parcount}

\DeclareRobustCommand{\parhead}[1]{\textbf{#1}~}

\usepackage[usenames,dvipsnames]{xcolor}
\definecolor{shadecolor}{gray}{0.9}

\newcommand{\red}[1]{\textcolor{BrickRed}{#1}}

\newcommand{\green}[1]{\textcolor{OliveGreen}{#1}}
\newcommand{\blue}[1]{\textcolor{MidnightBlue}{#1}}

\usepackage{graphicx}
\usepackage[labelfont=bf]{caption}
\usepackage[format=hang]{subcaption}

\usepackage{booktabs, array}

\usepackage{natbib}
\usepackage[algoruled]{algorithm2e}
\usepackage{listings}
\usepackage{fancyvrb}
\fvset{fontsize=\small}
\usepackage[colorlinks,linktoc=all]{hyperref}
\usepackage[all]{hypcap}
\hypersetup{citecolor=MidnightBlue}
\hypersetup{linkcolor=MidnightBlue}
\hypersetup{urlcolor=MidnightBlue}

\usepackage
[acronym,smallcaps,nowarn,section,nonumberlist]{glossaries}
\glsdisablehyper{}

\lstdefinestyle{mystyle}{
    commentstyle=\color{OliveGreen},
    numberstyle=\tiny\color{black!60},
    stringstyle=\color{BrickRed},
    basicstyle=\ttfamily\scriptsize,
    breakatwhitespace=false,
    breaklines=true,
    captionpos=b,
    keepspaces=true,
    numbers=none,
    numbersep=5pt,
    showspaces=false,
    showstringspaces=false,
    showtabs=false,
    tabsize=2
}
\RequirePackage[bf,small]{titlesec}

\usepackage{iclr2017_conference, times,}
\newacronym{ELBO}{elbo}{Evidence Lower Bound}
\newacronym{VI}{vi}{variational inference}
\newacronym{KL}{kl}{Kullback-Leibler}
\newacronym{RNN}{rnn}{recurrent neural network}
\newacronym{RNNs}{rnns}{recurrent neural networks}
\newacronym{RNNLM}{rnnlm}{Recurrent Neural Network-based language model}
\newacronym{TopicRNN}{TopicRNN}{TopicRNN}
\newacronym{TopicLSTM}{TopicLSTM}{TopicLSTM}
\newacronym{TopicGRU}{TopicGRU}{TopicGRU}
\newacronym{LSTM}{LSTM}{LSTM}
\newacronym{GRU}{GRU}{GRU}
\newacronym{IMDB}{IMDB}{IMDB}

\renewcommand{\gg}{\,\|\,}

\newcommand{\diag}{\textrm{diag}}

\newcommand{\EE}[2]{\mathbb{E}_{#1}\left[ #2 \right]}

 \iclrfinalcopy
\usepackage{hyperref}
\usepackage{url}
\usepackage{amssymb}

\title{TopicRNN: A Recurrent Neural Network \\with Long-Range Semantic Dependency}

\author{
Adji B. Dieng \thanks{Work was done while at Microsoft Research.} \\
Columbia University\\
\texttt{abd2141@columbia.edu} \\
\And\\
\And\\
\And\\
\And\\
\And\\
\And\\
\And
Chong Wang \\
Deep Learning Technology Center \\
Microsoft Research\\
\texttt{chowang@microsoft.com} \\
\And\\
\And\\
\And\\
\AND
Jianfeng Gao\\
Deep Learning Technology Center \\
Microsoft Research\\
\texttt{jfgao@microsoft.com} \\
\And\\
\And
John Paisley \\
Columbia University\\
\texttt{jpaisley@columbia.edu} 
}

\begin{document}

\maketitle

\begin{abstract}
  In this paper, we propose \acrlong{TopicRNN}, a \gls{RNN}-based 
  language model designed to directly capture 
      the global semantic meaning relating words in a document via latent topics.
  Because of their sequential nature, \acrshort{RNN}s are good at capturing 
  the local structure of a word sequence -- both 
  semantic and syntactic -- but might face
  difficulty remembering long-range dependencies. Intuitively, 
    these long-range dependencies 
    are of semantic nature.   In contrast, latent topic models
  are able to capture   the global semantic   structure of a document
      but do not account for word ordering.
  The proposed \acrlong{TopicRNN} model integrates the 
  merits of \acrshort{RNN}s and
  latent topic models:        it captures local (syntactic) dependencies using an \acrshort{RNN}
  and   global (semantic) dependencies using latent topics.
  Unlike previous work 
  on contextual \acrshort{RNN} language modeling,   our model is learned end-to-end. 
  Empirical results on   word prediction   show that 
  \acrlong{TopicRNN} outperforms existing contextual \acrshort{RNN} baselines.
    In addition, \acrlong{TopicRNN} can be used as an unsupervised
  feature extractor for documents. We do this for sentiment analysis
  on the IMDB movie review dataset and report an error rate of
  $6.28\%$. This is comparable to the state-of-the-art $5.91\%$
  resulting from a semi-supervised approach. Finally,
  \acrlong{TopicRNN} also yields sensible topics, making it a useful
  alternative to document models such as latent Dirichlet allocation.
      
\end{abstract}

\section{INTRODUCTION}\label{sec:introduction}
When reading a document, short or long, humans have a mechanism that
somehow allows them to remember the gist of what they have read so
far. Consider the following example:

``\textit{The \textcolor{blue}{U.S.}\textcolor{blue}{presidential}
  \textcolor{blue}{race} isn't only drawing attention and controversy
  in the \textcolor{blue}{United States} -- it's being closely watched
  across the globe. But what does the rest of the world think about a
  \textcolor{blue}{campaign} that has already thrown up one surprise
  after another?  CNN asked 10 journalists for their take on the
  \textcolor{blue}{race} so far, and what their  \textcolor{blue}{country} might be
hoping for in \textcolor{blue}{America}'s next
\textcolor{blue}{\bf{---}}}''

The missing word in the text above
is easily predicted by any human to be 
either \textit{President} or \textit{Commander in Chief} or their
synonyms. There have been various language models -- from simple $n$-grams
to the most recent \acrshort{RNN}-based language
models -- that aim to solve this problem of predicting correctly the
subsequent word in an observed sequence of words. 

A good language model should capture at least two important properties
of natural language.  The first one is correct syntax. 
In order to do prediction that enjoys this property, 
we often only need to consider a few preceding words. 
Therefore, correct syntax is more of a local property.
Word order matters in this case. 
The second property is the semantic coherence of the prediction. To achieve
this, we often need to consider many preceding words to understand the global semantic meaning of the sentence or document. 
The ordering of the words usually matters much less in this case.

Because they only consider a fixed-size context
window of preceding words,
traditional $n$-gram and neural probabilistic language
models~\citep{bengio2003neural} have difficulties in capturing
global semantic information. To overcome this, \acrshort{RNN}-based 
language models~\citep{mikolov2010recurrent,mikolov2011extensions} 
use hidden states to
``remember'' the history of a word sequence. However, 
none of these approaches explicitly model the two main 
properties of language mentioned above, correct syntax and semantic coherence. 
Previous work by \cite{chelba2000structured} and \cite{gao2004dependence}
exploit syntactic or semantic parsers to capture
long-range dependencies in language.

In this paper, we propose \acrlong{TopicRNN}, a \acrshort{RNN}-based
language model that is designed to directly capture long-range
semantic dependencies via latent topics. These topics provide context
to the \acrshort{RNN}.
 Contextual \acrshort{RNN}s have received 
a lot of attention~\citep{mikolov2012context, mikolov2014learning, ji2015document, lin2015hierarchical, ji2016latent, ghosh2016contextual}.
 However, the models
closest to ours are the contextual \acrshort{RNN} model proposed
by~\cite{mikolov2012context} and its most recent extension to
the long-short term memory (LSTM)
architecture~\citep{ghosh2016contextual}. These models use pre-trained
topic model features as an additional input to the hidden states
and/or the output of the \acrshort{RNN}. In contrast, \acrlong{TopicRNN} does
not require pre-trained topic model features and can be learned in an
end-to-end fashion. We introduce an automatic way for handling
stop words that topic models usually have difficulty dealing with.
Under a comparable model size set up, \acrlong{TopicRNN} achieves better
perplexity scores than the contextual \acrshort{RNN}
model of~\citet{mikolov2012context} on the Penn TreeBank dataset
\footnote{\citet{ghosh2016contextual} did not 
publish results on the PTB and we did not find the code online.}.
Moreover, \acrlong{TopicRNN} can be used as an unsupervised feature extractor
for downstream applications. For example, we derive document features
of the IMDB movie review dataset using \acrlong{TopicRNN} for
sentiment classification. We reported an error rate of $6.28\%$. This
is close to the state-of-the-art
$5.91\%$~\citep{miyato2016adversarial} despite that we do not use the
labels and adversarial training in the feature extraction stage.

The remainder of the paper is organized as follows: 
Section\nobreakspace \ref {sec:background} provides background on \acrshort{RNN}-based language models and
probabilistic topic models. Section\nobreakspace \ref {sec:TOPICRNN} 
describes the \acrlong{TopicRNN} network architecture,
its generative process and how to perform inference for it. 
Section\nobreakspace \ref {sec:experiments} presents per-word perplexity 
results on the Penn TreeBank dataset
and the classification error rate on the IMDB $100$K dataset.  
Finally, we conclude and provide future research directions in Section\nobreakspace \ref {sec:discussion}.

 \section{BACKGROUND}\label{sec:background}
We present the background 
necessary for building 
the \acrlong{TopicRNN} model. We first review 
\acrshort{RNN}-based language modeling, followed by a discussion 
on the construction of latent topic models. 

\subsection{Recurrent Neural Network-Based Language Models}

Language modeling is fundamental to many
applications. Examples include speech 
recognition and machine translation. 
A language model is a probability distribution 
over a sequence of words
in a predefined vocabulary. More formally, let $V$ be a
vocabulary set and $y_1, ..., y_T$ a sequence 
of $T$ words with each $y_t\in V$.
A language model measures the likelihood of a 
sequence through a joint probability distribution,
\begin{equation*}
p(y_1, ..., y_T) = p(y_1)\prod_{t=2}^{T} p(y_t | y_{1: t-1}).
\end{equation*}
Traditional $n$-gram and feed-forward neural network language
models~\citep{bengio2003neural} typically make 
Markov assumptions about the sequential
dependencies between words, where the chain rule shown 
above limits conditioning to a fixed-size context window.

\acrshort{RNN}-based language models~\citep{mikolov2011extensions} 
sidestep this Markov assumption by defining the 
conditional probability of each word $y_t$ given all the 
previous words $y_{1:t-1}$ 
through a hidden state $h_t$ (typically via a softmax function):
\begin{align*}
  p(y_t | y_{1: t-1}) &\triangleq p(y_t | h_t),  \\
  h_t & = f(h_{t-1}, x_t).
\end{align*}
The function $f(\cdot)$ can either be a 
standard \acrshort{RNN} cell or a more complex 
cell such as GRU~\citep{cho2014learning} 
or LSTM~\citep{hochreiter1997long}. The input and target words 
are related via the relation $x_t \equiv y_{t-1}$. 
These \acrshort{RNN}-based language models
have been quite successful~\citep{mikolov2011extensions,chelba2013one,jozefowicz2016exploring}.

While in principle \acrshort{RNN}-based models can ``remember''
arbitrarily long histories if provided enough capacity, in
practice such large-scale neural networks can easily encounter difficulties 
during optimization~\citep{bengio1994learning,pascanu2013difficulty,sutskever2013training} or
overfitting issues~\citep{srivastava2014dropout}. 
Finding better ways to model
long-range dependencies in language modeling is therefore 
an open research challenge. As motivated in the
introduction, much of the long-range dependency
in language comes from semantic coherence, not from syntactic structure 
which is
more of a local phenomenon. 
Therefore, models that can capture long-range
semantic dependencies in language are complementary
to \acrshort{RNN}s. In the following section, we describe 
a family of such models called probabilistic topic models.

\subsection{Probabilistic Topic Models}
Probabilistic topic models are a family of models 
that can be used to capture global
semantic coherency \citep{blei2009topic}. 
They provide a powerful tool for
summarizing, organizing, and navigating document collections. 
One basic goal of such models is to find groups of words that tend to
co-occur together in the same document. 
These groups of words are called topics and
represent a probability distribution 
that puts most of its mass on this subset of the vocabulary. 
Documents are then represented as 
mixtures over these latent topics. 
Through posterior inference, the learned topics capture the 
semantic coherence of the words they cluster 
together~\citep{mimno2011optimizing}.

The simplest topic model is latent Dirichlet allocation
(LDA)~\citep{blei2003latent}.  It assumes $K$ underlying topics $\beta
= \{\beta_1,\dots,\beta_K\}$ , each of which is a distribution over a fixed
vocabulary. The generative process of LDA is as follows:\\
First generate the $K$ topics, $\beta_k \sim_{iid} {\rm Dirichlet}(\tau)$. 
Then for each document containing words $y_{1:T}$, 
independently generate document-level variables and data:
\begin{enumerate}
  \item Draw a document-specific topic proportion vector $\theta\sim {\rm Dirichlet}(\alpha)$. 
  \item For the $t$th word in the document,
\begin{enumerate}
  \item Draw topic assignment $z_t \sim {\rm Discrete}(\theta)$. 
  \item Draw word $y_t \sim {\rm Discrete}(\beta_{z_t})$. 
\end{enumerate}
\end{enumerate}
Marginalizing each $z_t$, we obtain the probability of
$y_{1:T}$ via a matrix factorization followed 
by an integration over the latent variable $\theta$,
\begin{align}
  p(y_{1:T} |\beta) = \int p(\theta)\prod_{t=1}^T \sum_{z_t} p(z_t |\theta)p(y_t | z_t,
\beta) {\rm d}\theta =\int p(\theta) \prod_{t=1}^T (\beta \theta)_{y_t} {\rm d}\theta.\label{eq:lda-prob}
\end{align}
In LDA the prior distribution on the topic proportions
is a Dirichlet distribution; it can be replaced by many 
other distributions. For example, the correlated topic 
model~\citep{blei2006correlated} uses a log-normal distribution.
Most topic models are ``bag of words'' models in that 
word order is ignored. This makes it easier for topic models to
capture global semantic information. However, this is also one of
the reasons why topic models do not perform well on general-purpose
language modeling applications such as word prediction.
While bi-gram topic
models have been proposed~\citep{wallach2006topic}, 
higher order models quickly become intractable.

Another issue encountered by topic models is that they do not
model stop words well. This is because  stop words usually do not carry semantic meaning;  their appearance is mainly 
to make the sentence more readable according to the 
grammar of the language.
 They also appear frequently in almost every document 
 and can co-occur with almost
any word\footnote{\cite{wallach2009rethinking} described using
  asymmetric priors to alleviate this issue. Although it is not clear
how to use this idea in \acrlong{TopicRNN},  we plan to investigate such priors in future work.}.
 In practice, these stop words 
are chosen using tf-idf~\citep{blei2009topic}.

\section{The \acrlong{TopicRNN} Model}\label{sec:TOPICRNN}
\begin{figure*}[t]
  \begin{subfigure}{.47\textwidth}
\centering
\includegraphics[width=0.95\textwidth]{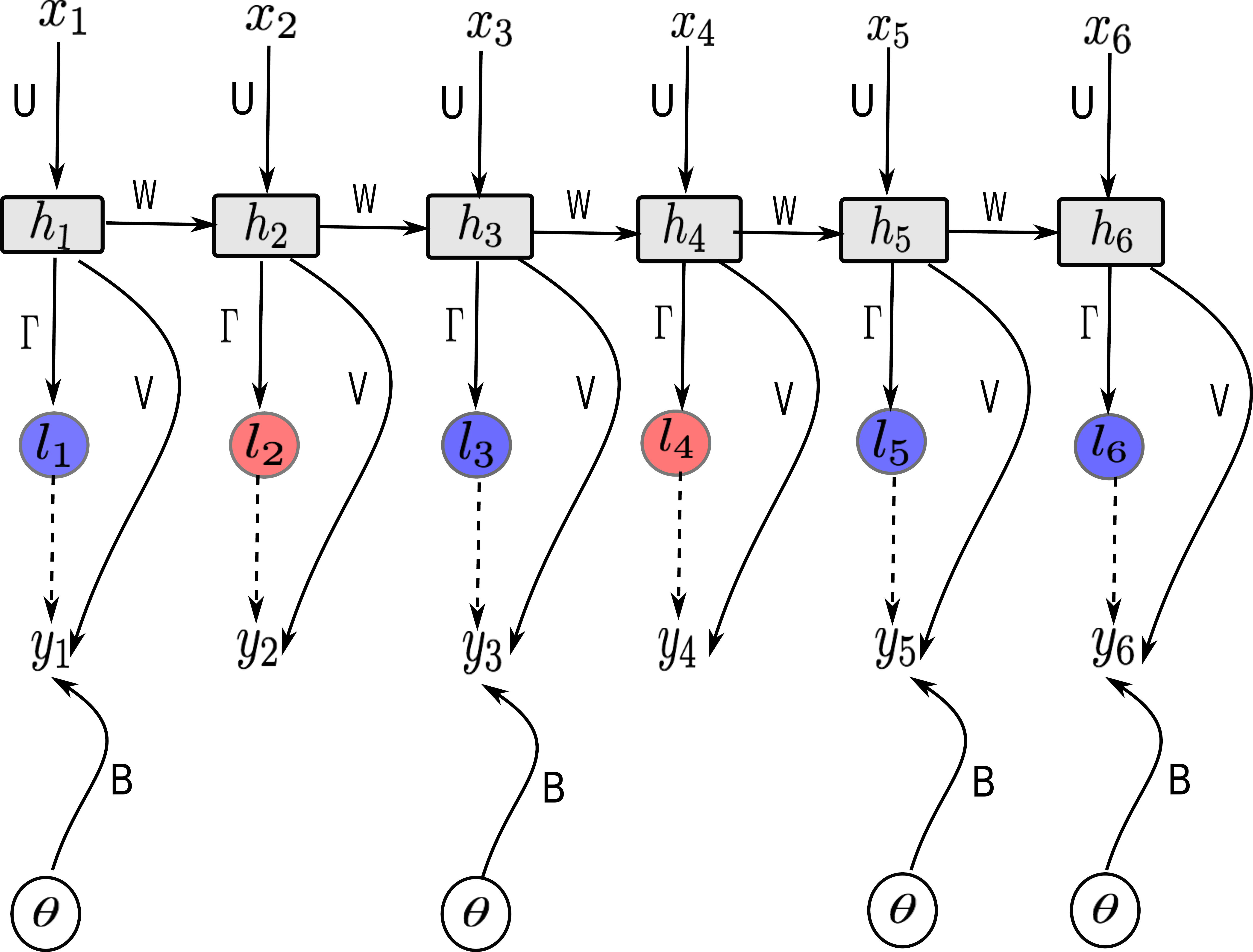}
\caption{}
\end{subfigure}
\begin{subfigure}{.56\textwidth}
\centering
\includegraphics[width=0.9\textwidth]{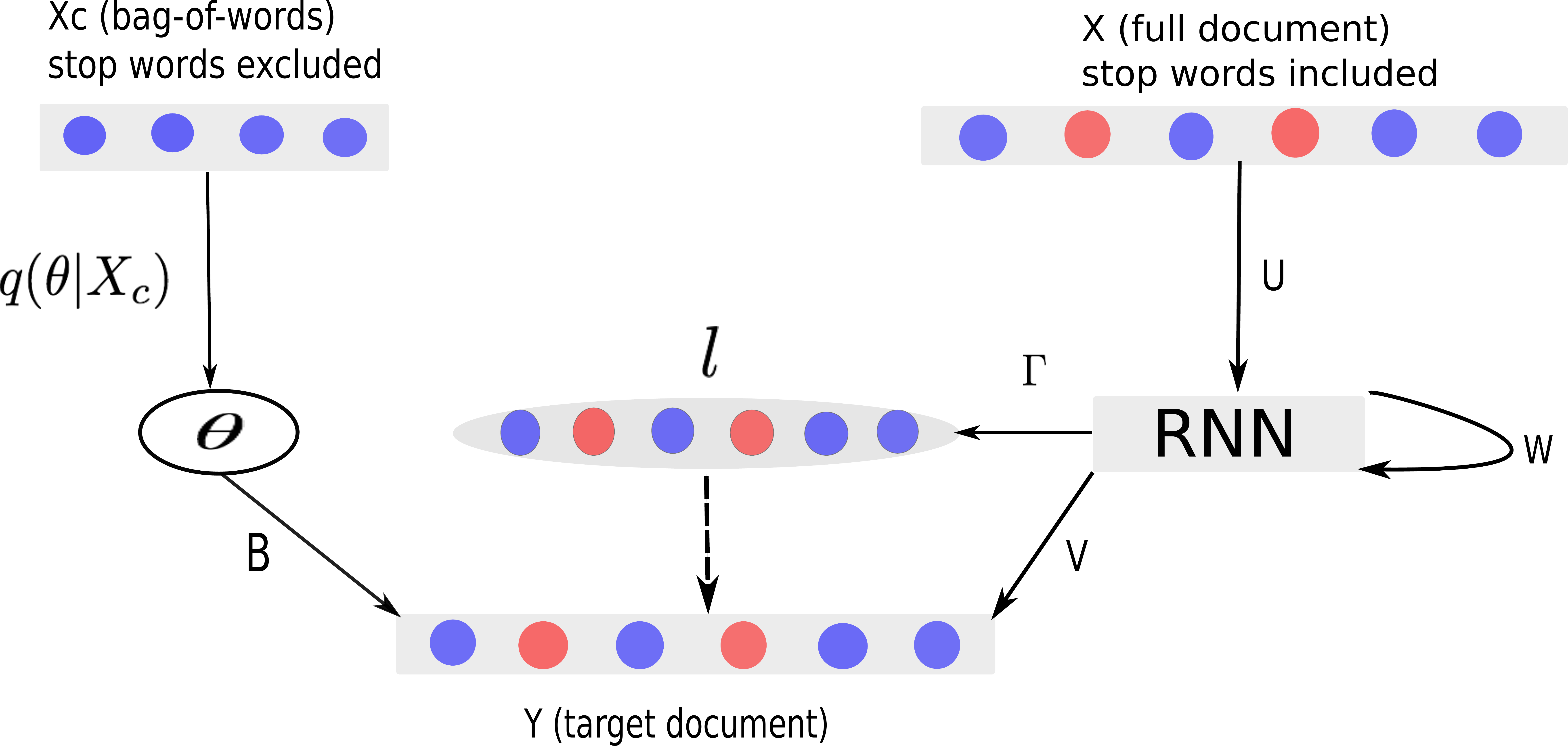}
\caption{}
  \end{subfigure}
  \caption{(a) The unrolled \acrlong{TopicRNN} architecture: $x_1, ..., x_6$
  	are words in the document, $h_t$ is the state of the \acrshort{RNN} at time step $t$, 
	$x_i \equiv y_{i-1}$, $l_1, ..., l_6$ are stop word indicators, and
	$\theta$ is the latent representation of the input document
    and is unshaded by convention. (b) The
     \acrlong{TopicRNN} model architecture in its compact form: 
     	$l$ is a binary vector that indicates whether each word in 
	the input document is a stop word or not. Here \red{red} indicates stop words
	and \blue{blue} indicates content words.}
\label{fig:architecture}
\end{figure*}

We next describe the proposed \acrlong{TopicRNN}
model. In \acrlong{TopicRNN}, latent topic models are used to capture
global semantic dependencies so that the \acrshort{RNN} can focus its modeling capacity
on the local dynamics of the sequences. With this joint
modeling, we hope to achieve better overall performance on downstream
applications.

\parhead{The model.} \acrlong{TopicRNN} is a generative model. For a document containing the words $y_{1:T}$, 
\begin{enumerate}
  \item Draw a topic vector\footnote{Instead of using the Dirichlet distribution,
    we choose the Gaussian distribution. This allows for more flexibility in
  the sequence prediction problem and also has advantages during inference.}
    $\theta\sim N(0,I)$. 
  \item Given word $y_{1:t-1}$, for the $t$th word $y_t$ in the document,
\begin{enumerate}
  \item Compute hidden state $h_t = f_W(x_t, h_{t-1})$, where we let $x_t \triangleq y_{t-1}$.  
  \item Draw stop word indicator $l_t \sim {\rm Bernoulli}(\sigma(\Gamma^\top h_t))$, with $\sigma$ the sigmoid function.
  \item Draw word $y_t \sim p(y_t | h_t, \theta, l_t,B)$, where
  \begin{align*}
  p(y_t = i | h_t, \theta, l_t,B) \propto \exp\left(v_i^\top h_t  + (1-l_t) b_i^\top\theta\right).
\end{align*}
\end{enumerate}
\end{enumerate}

The stop word indicator $l_t$ controls how the topic vector $\theta$ affects the output. 
If $l_t = 1$ (indicating $y_t$ is a stop word), the topic vector
$\theta$ has no contribution to the output. Otherwise, we add a bias
to favor those words that are more likely to appear when 
mixing with $\theta$, as measured by the dot product 
between $\theta$ and the latent word vector $b_i$ for the $i$th vocabulary word.  
As we can see, the long-range semantic
information captured by $\theta$ directly
affects the output through an additive procedure. Unlike~\citet{mikolov2012context}, the contextual
information is not passed to the hidden layer of the \acrshort{RNN}.
The main reason behind our choice of using the topic vector as bias
instead of passing it into the hidden states of the RNN is because
it enables us to have a clear separation of the contributions of global
semantics and those of local dynamics. The global semantics come from
the topics which are meaningful when stop words are excluded. 
However these stop words are needed for the local dynamics of the language
model. We hence achieve this separation of global vs local via a
binary decision model for the stop words.
It is unclear how to achieve this if we pass the topics to the hidden
states of the RNN. This is because the hidden states of the RNN will
account for all words (including stop words) whereas the topics
exclude stop words.  

We show the unrolled graphical representation of \acrlong{TopicRNN} in
Figure\nobreakspace \ref {fig:architecture}(a). 
We denote all model parameters 
as $\Theta = \{\Gamma, V, B, W, W_c\}$ (see Appendix\nobreakspace \ref {sec:appendix_params}
for more details). Parameter $W_c$ is for the inference network, which
we will introduce below.
The observations are the word sequences $y_{1:T}$ and stop word
indicators $l_{1:T}$.\footnote{Stop words can be determined using one of 
the several lists available online. For example, 
\url{http://www.lextek.com/manuals/onix/stopwords2.html}}
The log marginal likelihood of the sequence
$y_{1:T}$ is
\begin{align}
  \log p(y_{1:T}, l_{1:T}|h_t) = \log \int p(\theta) \prod_{t=1}^T p(y_t | h_t, l_t, \theta) p(l_t
  | h_t) {\rm d} \theta. \label{eq:likelihood}
\end{align}

\parhead{Model inference.} Direct optimization of
Equation\nobreakspace \textup {\ref {eq:likelihood}} is intractable so we use variational
inference for approximating this 
marginal~\citep{jordan1999introduction}. Let $q(\theta)$ be the
variational distribution on the marginalized variable $\theta$. 
We construct the variational objective function, also called 
the evidence lower bound (\acrshort{ELBO}), as follows:
\begin{align*}
  \mathcal{L}(y_{1:T}, l_{1:T}|q(\theta), \Theta) &\triangleq~
  \EE{q(\theta)} {\sum_{t=1}^T  \log p(y_t | h_t, l_t, \theta) + \log p(l_t
  | h_t) + \log p(\theta) - \log q(\theta)} \\ 
  & \leq ~\log p(y_{1:T}, l_{1:T} |h_t,\Theta).
\end{align*}
Following the proposed variational autoencoder 
technique, we choose the form of
$q(\theta)$ to be an inference network using a feed-forward neural
network~\citep{kingma2013auto,miao2015neural}. 
Let $X_c\in \mathcal{N}_+^{|V_c|}$
be the term-frequency representation of $y_{1:T}$ {\it excluding} 
stop words (with $V_c$ the vocabulary size without the
stop words). The variational autoencoder inference 
network $q(\theta|X_c, W_c)$ with parameter
$W_c$ is a feed-forward neural network with ReLU activation units that
projects $X_c$ into a $K$-dimensional latent space. Specifically,  we
have
\begin{align*}
	q(\theta | X_c, W_c) &= N(\theta; \mu(X_c), \diag(\sigma^2(X_c))), \\
  \mu(X_c) &= W_1 g(X_c) + a_1,\\
	\log\sigma(X_c) &= W_2 g(X_c) + a_2,
\end{align*}
where $g(\cdot)$ denotes the feed-forward neural network. The weight
matrices $W_1$, $W_2$ and biases $a_1$, $a_2$ are shared across
documents. Each document has its own $\mu(X_c)$ and $\sigma(X_c)$
resulting in a unique distribution $q(\theta | X_c)$ for each
document.  The output of the inference network is a
distribution on $\theta$, which we regard as the
summarization of the semantic information, similar to the topic
proportions in latent topic models. We show the role of the inference
network in Figure\nobreakspace \ref {fig:architecture}(b).  During training, the
parameters of the inference network and the model are jointly learned
and updated via truncated backpropagation through time using the
Adam algorithm~\citep{kingma2014adam}. We use stochastic samples from
$q(\theta|X_c)$ and the reparameterization
trick towards this end~\citep{kingma2013auto,rezende2014stochastic}.

\parhead{Generating sequential text and computing perplexity.}
Suppose we are given a word sequence $y_{1:t-1}$, from which we have
an initial estimation of $q(\theta | X_c)$. To generate the next word
$y_t$, we compute the probability distribution of $y_t$ given
$y_{1:t-1}$ in an online fashion. 
We choose
$\theta$ to be a point estimate $\hat{\theta}$, the mean of its 
current distribution $q(\theta | X_c)$. 
Marginalizing over the 
stop word indicator $l_t$ which is unknown prior to 
observing $y_t$, the approximate distribution of $y_t$ is 
\begin{align*}
  p(y_t | y_{1:t-1}) \approx  \sum_{l_t} p(y_t |
  h_t,\hat{\theta}, l_t)p(l_t | h_t) .
\end{align*}
The predicted word $y_t$ is a sample from this predictive distribution.
We update $q(\theta | X_c)$ by including $y_t$ to $X_c$ if $y_t$ is 
not a stop word. However, updating $q(\theta| X_c)$ after each word
prediction is expensive, so we use a sliding window
as was done in~\citet{mikolov2012context}. To compute the perplexity, we 
use the approximate predictive distribution above.

\parhead{Model Complexity.}
\acrlong{TopicRNN} has a complexity of 
$O(H\times H + H \times (C + K) + W_c)$,
where $H$ is the size of the hidden layer of the \acrshort{RNN},
$C$ is the vocabulary size, $K$ is the dimension of the 
topic vector, and $W_c$ is the number of parameters 
of the inference network. 
The contextual 
\acrshort{RNN} of~\citet{mikolov2012context} 
accounts for $O(H\times H + H \times (C + K))$, not including the pre-training process, which
might require more parameters than the additional $W_c$ in our complexity. 
 
\section{EXPERIMENTS}\label{sec:experiments}
We assess the performance of our proposed \acrlong{TopicRNN} model on  
word prediction and sentiment analysis\footnote{Our code will be made publicly available for reproducibility.}. 
For word prediction we use the Penn TreeBank 
dataset, a standard benchmark for assessing 
new language models~\citep{marcus1993building}. 
For sentiment analysis we use the IMDB 100k 
dataset~\citep{maas2011learning}, also a 
common benchmark dataset for this application\footnote{These datasets are publicly available 
at \url{http://www.fit.vutbr.cz/~imikolov/rnnlm/simple-examples.tgz} 
and \url{http://ai.stanford.edu/~amaas/data/sentiment/}.}.
We use \acrshort{RNN}, LSTM, and GRU cells
in our experiments leading to \acrlong{TopicRNN}, 
\acrlong{TopicLSTM}, and \acrlong{TopicGRU}.
\begin{table}[!hbpt]
\caption{\label{tab:topics} Five Topics from 
the \acrlong{TopicRNN} Model with 100 Neurons and 
50 Topics on the PTB Data. (The word \textit{s{\&}p} 
below shows as \textit{sp} in the data.)}
\begin{center}
\resizebox{0.7\columnwidth}{!}{
\begin{tabular}{c|c|c|c|c}
\toprule
Law & Company& Parties & Trading & Cars\\ \midrule
law  & spending & democratic & stock & gm \\
lawyers & sales & republicans & s{\&}p & auto \\
judge & advertising & gop & price & ford \\
rights & employees & republican & investor & jaguar \\
attorney & state & senate & standard & car \\
court & taxes & oakland & chairman & cars \\
general & fiscal & highway & investors & headquarters \\
common & appropriation & democrats & retirement & british \\
mr  & budget & bill & holders &executives \\
insurance & ad & district & merrill & model \\\bottomrule
\end{tabular}}
\end{center}
\end{table}
\begin{figure*}[!hbpt]
   \centering
    {\includegraphics[scale=0.22]{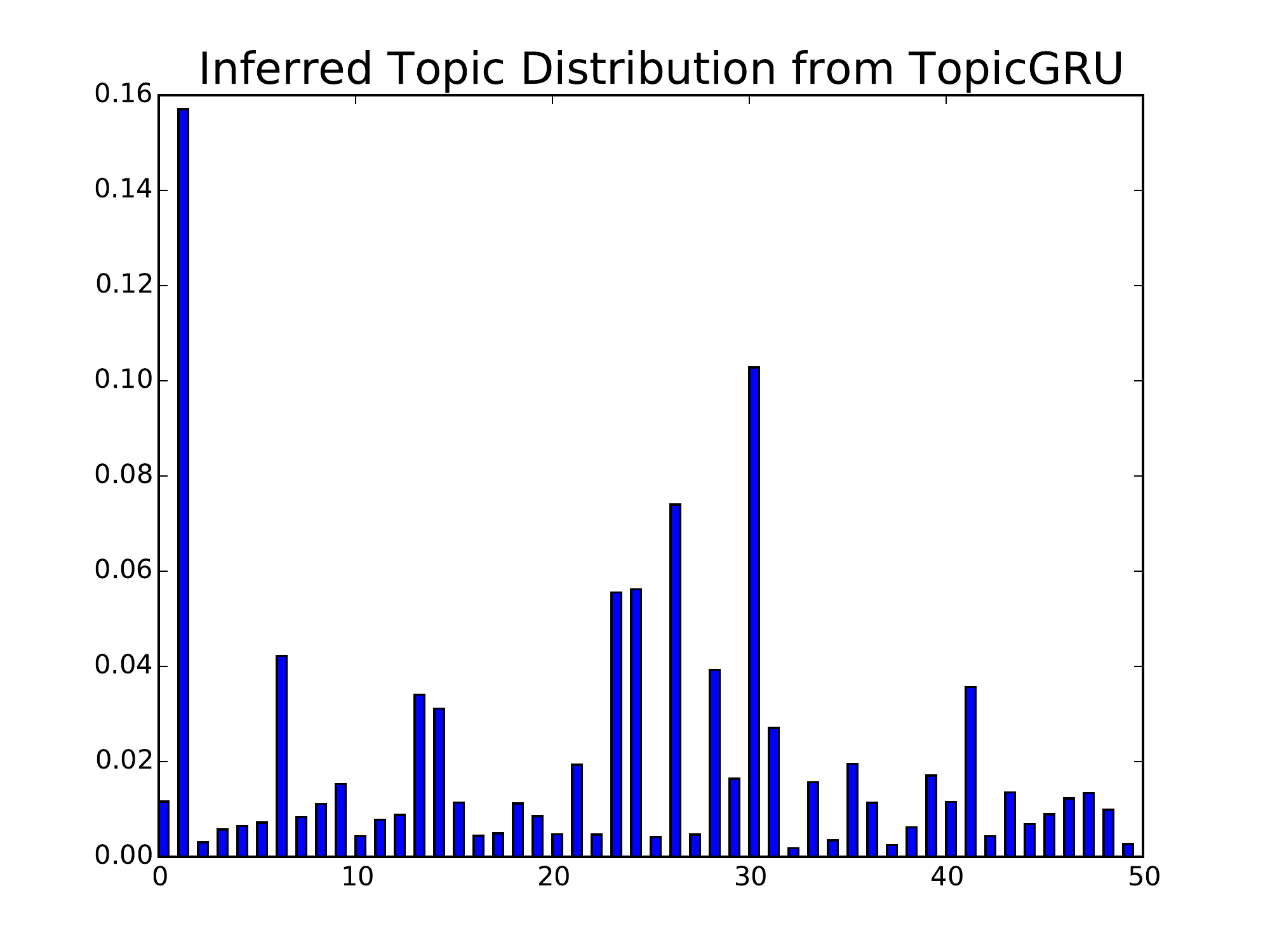}}
    {\includegraphics[scale=0.22]{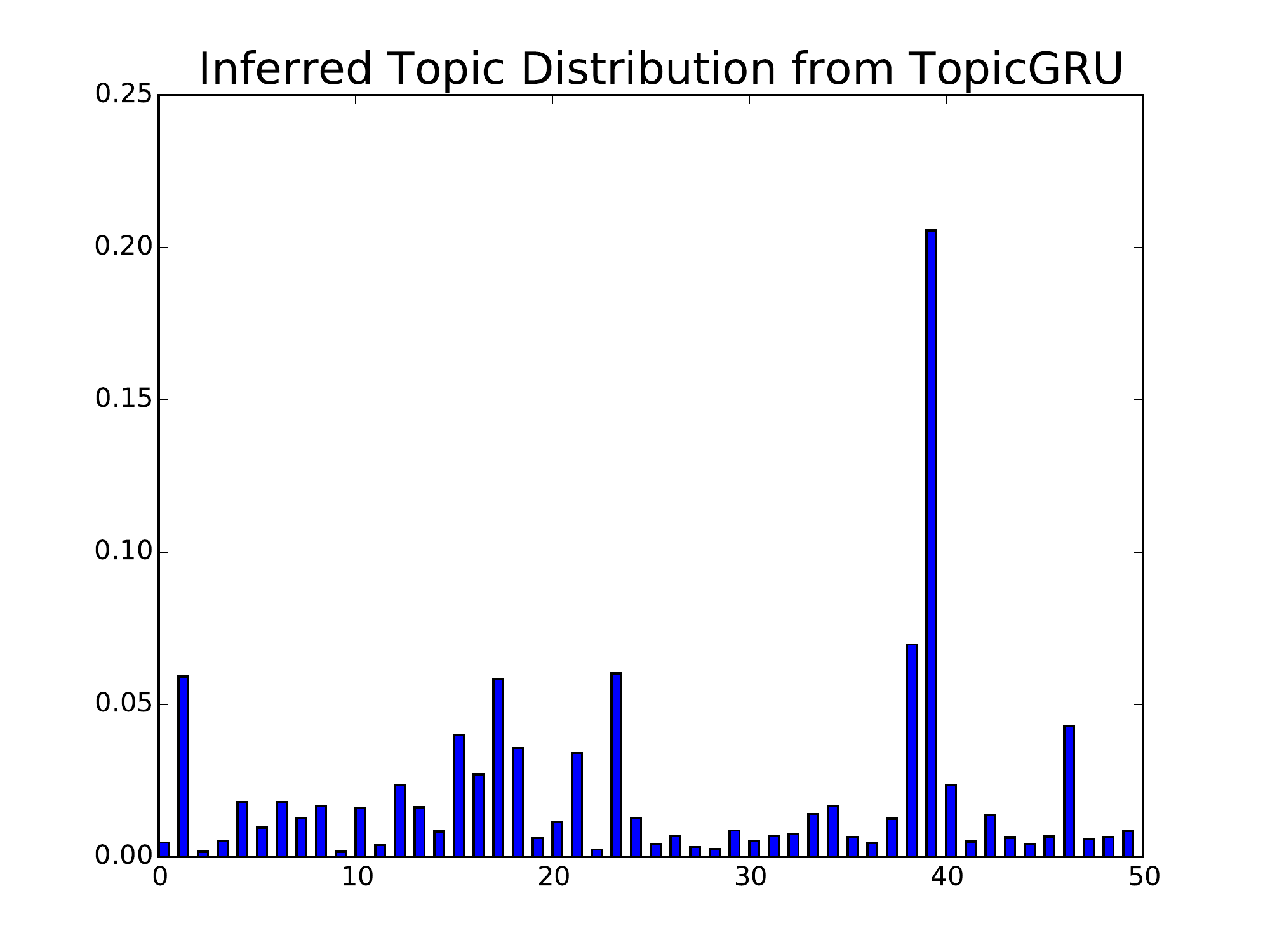}} 
    {\includegraphics[scale=0.22]{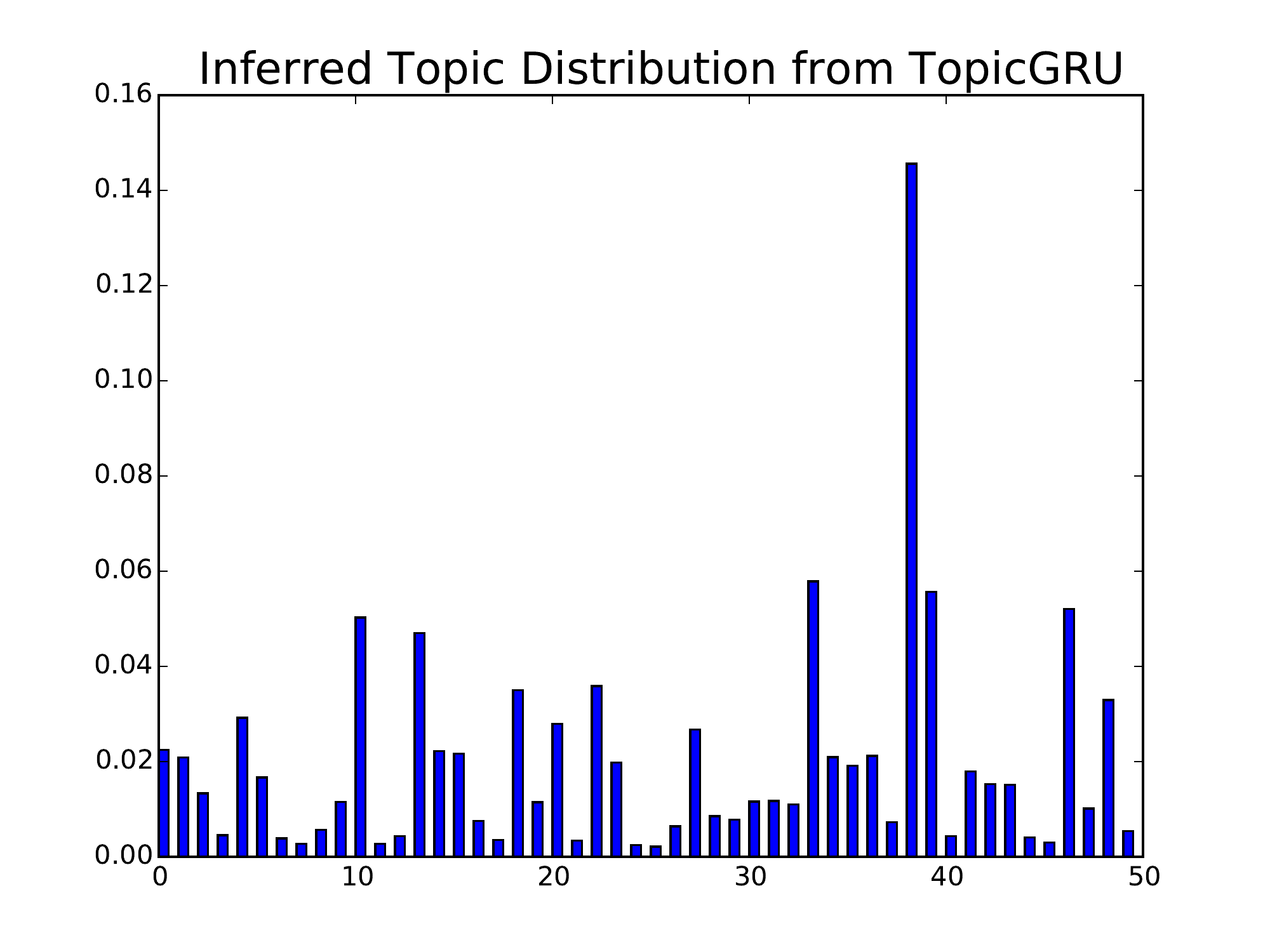}}
    \caption{Inferred distributions using TopicGRU on three different documents.
    The content of these documents is added on the appendix.
    This shows that some of the topics are being picked up
    depending on the input document.}
    \label{fig:theta}
\end{figure*}
\begin{table}[!hbpt]
    \caption{\acrlong{TopicRNN} and its counterparts exhibit lower perplexity scores across 
    different network sizes than reported in ~\citet{mikolov2012context}. 
    \label{tab:pp}
    Table\nobreakspace \ref {tab:a} shows per-word perplexity scores for 10 neurons. 
    Table\nobreakspace \ref {tab:b} and Table\nobreakspace \ref {tab:c} correspond to per-word 
    perplexity scores for 100 and 300 neurons respectively.
    These results prove TopicRNN has more generalization capabilities:
    for example we only need a TopicGRU with 100 neurons 
    to achieve a better perplexity than stacking 2 LSTMs with 
    200 neurons each: 112.4 vs 115.9)}
     \centering
    \begin{subtable}{0.5\columnwidth}
        \caption{}\label{tab:a}
       \begin{tabular}{lcc}
\toprule
10 Neurons	&Valid	&Test \\\midrule
\acrshort{RNN} (no features) &$239.2$&$225.0$ \\
\acrshort{RNN} (LDA features)	 &$197.3$	 &$187.4$ \\
TopicRNN	 &$184.5$	 & $172.2$\\
TopicLSTM & $188.0$ & $175.0$\\
TopicGRU & $178.3$ & \bf{166.7} \\\bottomrule
\end{tabular}
    \end{subtable}
    \hspace{5pt}
    \begin{subtable}{0.4\columnwidth}
        \caption{}\label{tab:b}
        \begin{tabular}{lcc}
\toprule
100 Neurons & Valid & Test \\\midrule
\acrshort{RNN} (no features)  & $150.1$ & $142.1$ \\
\acrshort{RNN} (LDA features) & $132.3$ & $126.4$ \\
TopicRNN	 &$128.5$	 & $122.3$\\
TopicLSTM & $126.0$ & $118.1$\\
TopicGRU & $118.3$ & \bf{112.4} \\\bottomrule
\end{tabular}
    \end{subtable} 
     \centering
     \begin{subtable}{.5\columnwidth}
      \centering
        \caption{}\label{tab:c}
        \begin{tabular}{lcc}
\toprule
300 Neurons & Valid & Test \\\midrule
\acrshort{RNN} (no features)  & $ - $ & $124.7$ \\
\acrshort{RNN} (LDA features) & $ - $ & $113.7$ \\
TopicRNN	 &$118.3$	 &$112.2$\\
TopicLSTM & $104.1$ & $99.5$\\
TopicGRU & $99.6$ & \bf{97.3} \\\bottomrule
\end{tabular}
    \end{subtable} 
\end{table}
\subsection{Word Prediction}
We first tested \acrlong{TopicRNN} on the word prediction 
task using the Penn Treebank (PTB) 
portion of the Wall Street Journal. We use the 
standard split, where sections 0-20 (930K tokens) 
are used for training, sections 21-22 (74K tokens) for 
validation, and 
sections 23-24 (82K tokens) for 
testing~\citep{mikolov2010recurrent}. 
We use a vocabulary of size $10K$ that includes 
the special token \textit{unk} for 
rare words and \textit{eos} that indicates the 
end of a sentence. \acrlong{TopicRNN} takes documents 
as inputs. We split the PTB data into blocks 
of 10 sentences to constitute documents 
as done by~\citep{mikolov2012context}. 
The inference network takes as input the bag-of-words 
representation of the input document. 
For that reason, the vocabulary size of the inference network is reduced 
to $9551$ after excluding $449$ pre-defined stop words. 

In order to compare with previous work on 
contextual \acrshort{RNN}s we trained \acrlong{TopicRNN}
 using different network sizes. We performed 
 word prediction using a recurrent neural 
 network with 10 neurons, 100 neurons and 300 
 neurons. For these experiments, we used 
 a multilayer perceptron with 2 hidden layers 
 and 200 hidden units per layer for the inference network. 
 The number of topics was tuned depending on 
 the size of the \acrshort{RNN}. For 10 neurons 
 we used 18 topics. For 100 and 300 neurons we found 50 
 topics to be optimal. We used the validation set 
 to tune the hyperparameters of the model.
 We used a maximum of 15 epochs for the experiments 
 and performed early stopping 
 using the validation set. For comparison purposes 
 we did not apply dropout and used 1 layer 
 for the \acrshort{RNN} and its counterparts in all 
 the word prediction experiments as reported in Table\nobreakspace \ref {tab:pp}.
One epoch for 10 neurons takes $2.5$ minutes.
For 100 neurons, one epoch is completed in less than 
4 minutes. Finally, for 300 neurons one epoch takes 
less than 6 minutes. These experiments were ran
on Microsoft Azure NC12 that has 12 cores, 2 Tesla K80 GPUs,
and 112 GB memory.
First, we show five randomly drawn topics in Table\nobreakspace \ref {tab:topics}. These
results correspond to a network with 100 neurons.  We also illustrate
some inferred topic distributions for several documents from TopicGRU
in Figure\nobreakspace \ref {fig:theta}. Similar to standard topic models, these
distributions are also relatively peaky.
 
Next, we compare the performance of \acrlong{TopicRNN}
to our baseline contextual \acrshort{RNN}
using perplexity.  Perplexity can be thought of 
as a measure of surprise for a language model. It is 
defined as the exponential of the 
average negative log likelihood. Table\nobreakspace \ref {tab:pp} summarizes 
the results for different network sizes. 
We learn three things from these tables. First, 
the perplexity is reduced the larger 
the network size. Second, \acrshort{RNN}s with context 
features perform better 
than \acrshort{RNN}s without context features. Third, we 
see that \acrlong{TopicRNN} 
gives lower perplexity than the previous baseline result 
reported by~\cite{mikolov2012context}. Note that
to compute these perplexity scores for word prediction
we use a sliding window to compute $\theta$ as we move
along the sequences. The topic vector $\theta$ that is used 
from the current batch of words is estimated from 
the previous batch of words. This enables fair comparison 
to previously reported results \citep{mikolov2012context}.\footnote{We
adjusted the scores in Table\nobreakspace \ref {tab:pp} from what was previously reported
after correcting a bug in the computation of the ELBO.}

Another aspect of the \acrlong{TopicRNN} 
model we studied is its capacity to generate coherent text. 
To do this, we randomly drew a document from the test set
and used this document as seed input to the inference network 
to compute $\theta$. Our expectation is that 
the topics contained in this seed document 
are reflected in the generated text.
Table\nobreakspace \ref {tab:text} shows generated text from models 
learned on the PTB and IMDB datasets. See Appendix\nobreakspace \ref {sec:appendix_text} 
for more examples. 
\begin{table}[t]
\caption{\label{tab:text}Generated text using the \acrlong{TopicRNN} model 
on the PTB (top) and IMDB (bottom).}
\begin{center}
\begin{tabular}{l}
\begin{minipage}[t]{0.9\columnwidth}\textit{they believe that they had senior damages 
to guarantee and frustration of unk 
stations eos the rush to minimum effect in composite 
trading the compound base inflated rate 
before the common charter 's report eos wells 
fargo inc. unk of state control funds without openly 
scheduling the university 's exchange rate has been 
downgraded it 's unk said eos the united 
cancer \& began critical increasing rate of 
N N at N N to N N are less for the country 
to trade rate for more than three months \$ N workers were mixed eos}\vspace{3pt}
\end{minipage}\tabularnewline
\midrule\vspace{3pt}
\begin{minipage}[t]{0.9\columnwidth}\textit{lee is head to be watched unk month she 
eos but the acting surprisingly 
nothing is very good eos i cant believe that he 
can unk to a role eos may appear of 
for the stupid killer really to help with unk unk unk 
if you wan na go to it fell to the plot 
clearly eos it gets clear of this movie $70$ are 
so bad mexico direction regarding 
those films eos then go as unk 's walk and after unk 
to see him try to unk before that unk with this film}
\end{minipage}\tabularnewline
\end{tabular}
\end{center}
\end{table}

\subsection{Sentiment Analysis}
\begin{table}[t]
\caption{Classification error rate on IMDB 100k dataset. 
\acrlong{TopicRNN} provides the state of the art error rate on this dataset.} 
\label{tab:imdb}
\centering
\resizebox{0.8\columnwidth}{!}{\begin{tabular}{lr}
\toprule
Model & Reported Error rate\\\midrule
BoW (bnc) (Maas et al., 2011) & $12.20\%$\\
BoW ($b\Delta$ t\'c) (Maas et al., 2011) & $11.77\%$\\
LDA (Maas et al., 2011) & $32.58\%$\\
Full +  BoW (Maas et al., 2011) & $11.67\%$ \\
Full + Unlabelled + BoW (Maas et al., 2011) & $11.11\%$ \\
WRRBM  (Dahl et al., 2012) & $12.58\%$ \\
WRRBM + BoW (bnc) (Dahl et al., 2012) & $10.77\%$ \\
MNB-uni (Wang \& Manning, 2012) & $16.45\%$ \\
MNB-bi (Wang \& Manning, 2012) & $13.41\%$ \\
SVM-uni (Wang \& Manning, 2012) & $13.05\%$ \\
SVM-bi (Wang \& Manning, 2012) & $10.84\%$ \\
NBSVM-uni (Wang \& Manning, 2012) & $11.71\%$ \\
seq2-bown-CNN (Johnson \& Zhang, 2014) & $14.70\%$ \\
NBSVM-bi (Wang \& Manning, 2012) & $8.78\%$ \\
Paragraph Vector (Le \& Mikolov, 2014) & $7.42\%$ \\
SA-LSTM  with joint training (Dai \& Le, 2015) & $14.70\%$ \\
LSTM with tuning and dropout (Dai \& Le, 2015)  & $13.50\%$ \\
LSTM initialized with word2vec embeddings (Dai \& Le, 2015) & $10.00\%$ \\
SA-LSTM  with linear gain (Dai \& Le, 2015) & $9.17\%$ \\
LM-TM  (Dai \& Le, 2015) & $7.64\%$ \\
SA-LSTM (Dai \& Le, 2015)  & $7.24\%$ \\
\textbf{Virtual Adversarial (Miyato et al. 2016)} & \textbf{5.91\%}\\
\midrule
\textbf{TopicRNN } & \textbf{6.28\%} \\\bottomrule
\end{tabular}}
\end{table}
\begin{figure}[t]
\centering
\includegraphics[width=.95\textwidth]{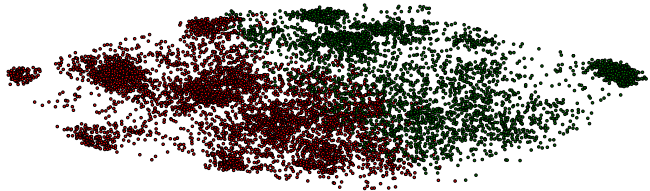}
\caption{Clusters of a sample of $10000$ movie reviews from the IMDB $100$K 
dataset using \acrlong{TopicRNN} as feature extractor. We used K-Means to cluster
 the feature vectors. We then used PCA to reduce 
 the dimension to two for visualization purposes.
 \red{red} is a negative review and \green{green} is a positive review.}
\label{fig:clusters}
\end{figure}
We performed sentiment analysis using \acrlong{TopicRNN} 
as a feature extractor on 
the IMDB 100K dataset. This data consists 
of 100,000 movie reviews from 
the Internet Movie Database (IMDB) website. The 
data is split into $75\%$ for training and $25\%$ for testing. 
Among the 75K training reviews, 50K are unlabelled 
and 25K are labelled as carrying either a positive or a negative sentiment. 
All 25K test reviews are labelled. We trained 
\acrlong{TopicRNN} on 65K random training reviews and used 
the remaining 10K reviews for validation. To learn a classifier, we passed 
the 25K labelled training reviews through the learned \acrlong{TopicRNN} model. We 
then concatenated the output of the inference network and the last 
state of the \acrshort{RNN} for each of these 25K reviews to compute the feature vectors. 
We then used these feature vectors to train a neural network with one hidden 
layer, 50 hidden units, and a sigmoid activation function to predict sentiment, exactly as 
 done in~\citet{le2014distributed}. 
 
To train the \acrlong{TopicRNN} model, we used a vocabulary 
of size 5,000 and mapped all other words to the \textit{unk} token. 
We took out 439 stop 
words to create the input of the inference network. 
We used 500 units and 2 layers 
for the inference network, and used 2 layers 
and 300 units per-layer for the \acrshort{RNN}. 
We chose a step size of 5 and defined 200 topics. 
We did not use any regularization such as dropout. 
We trained the model for 13 epochs and used the 
validation set to tune the 
hyperparameters of the model and track perplexity 
for early stopping. This experiment took close to 78 hours
on a MacBook pro quad-core with 16GHz of RAM. 
See Appendix\nobreakspace \ref {sec:appendix_imdb_topics} for the 
visualization of some of the topics learned from this data.

Table\nobreakspace \ref {tab:imdb} summarizes sentiment classification results from
\acrlong{TopicRNN} and other methods.  Our error rate is
$6.28\%$.\footnote{The experiments were solely based on TopicRNN.
Experiments using TopicGRU/TopicLSTM are being carried out and will be
added as an extended version of this paper.}  This is close to the
state-of-the-art $5.91\%$~\citep{miyato2016adversarial} despite that
we do not use the labels and adversarial training in the feature
extraction stage.  Our approach is most similar
to~\citet{le2014distributed}, where the features were extracted in a
unsupervised way and then a one-layer neural net was trained for
classification.

Figure\nobreakspace \ref {fig:clusters} shows the ability of \acrlong{TopicRNN} to cluster documents 
using the feature vectors as created 
during the sentiment analysis task. 
Reviews with positive sentiment are coloured 
in green while reviews carrying 
negative sentiment are shown in red. This shows that 
\acrlong{TopicRNN} can be used as an unsupervised feature 
extractor for downstream applications.
Table\nobreakspace \ref {tab:text} shows generated text from models 
learned on the PTB and IMDB datasets. See Appendix\nobreakspace \ref {sec:appendix_text}
for more examples. The overall generated text from IMDB encodes a negative 
sentiment.
 
\section{DISCUSSION AND FUTURE WORK}\label{sec:discussion}
In this paper we introduced \acrlong{TopicRNN}, 
a \acrshort{RNN}-based language model that combines 
\acrshort{RNN}s and latent topics to capture
local (syntactic) and global (semantic) dependencies 
between words.
The global dependencies as captured by the latent topics
serve as contextual bias to 
an \acrshort{RNN}-based language model.
This contextual information is learned jointly with the 
\acrshort{RNN} parameters by maximizing the evidence lower bound 
of variational inference. \acrlong{TopicRNN} yields competitive 
per-word perplexity on the Penn Treebank dataset 
compared to previous contextual \acrshort{RNN} models. We have 
reported a competitive classification error 
rate for sentiment analysis on the IMDB 100K dataset.  
 We have also illustrated the capacity of 
\acrlong{TopicRNN} to generate sensible topics and text. \\
In future work, we will study the performance of \acrlong{TopicRNN} when stop words 
are dynamically discovered during training. We will also extend \acrlong{TopicRNN} 
to other applications where capturing context is important such as in dialog modeling. 
If successful, this will allow us to have a model that performs 
well across different natural language processing applications. 
 
\bibliographystyle{abbrvnat}
\bibliography{main}

\appendix

\section{APPENDIX}\label{sec:appendix}
\subsection{Dimension of the parameters of the model:}\label{sec:appendix_params}
We use the following notation: C is the vocabulary size (including stop words),
H is the number of hidden units of the \gls{RNN}, K is the number of topics,
and E is the dimension of the inference network hidden layer.
Table\nobreakspace \ref {tab:params} gives the dimension of each of the parameters of the 
\acrlong{TopicRNN} model (ignoring the biases).
\begin{table}[!hbpt]
\caption{\label{tab:params} Dimensions of the parameters of the model.}
\begin{center}
\begin{tabular}{c|c|c|c|c|c|c|c|c}
 & U & $\Gamma$ & W & V & B & $\theta$ & $W_1$ & $W_2$\\ \hline
dimension  & C x H & H & H x H & H x C & K x C & K & E & E \\
\end{tabular}
\end{center}
\end{table}

\subsection{Documents used to infer the distributions on Figure\nobreakspace \ref {fig:theta}}\label{sec:appendix_docs}
Figure on the left: \blue{\textit{
'the', 'market', 'has', 'grown', 'relatively', 'quiet', 'since', 'the', 'china', 'crisis', 'but', 'if', 'the', 'japanese', 'return', 'in', 'force', 'their', 'financial', 'might', 'could', 'compensate', 'to', 'some', 'extent', 'for', 'local', 'investors', "'", '<unk>', 'commitment', 'another', 'and', 'critical', 'factor', 'is', 'the', 'u.s.', 'hong', 'kong', "'s", 'biggest', 'export', 'market', 'even', 'before', 'the', 'china', 'crisis', 'weak', 'u.s.', 'demand', 'was', 'slowing', 'local', 'economic', 'growth', '<unk>', 'strong', 'consumer', 'spending', 'in', 'the', 'u.s.', 'two', 'years', 'ago', 'helped', '<unk>', 'the', 'local', 'economy', 'at', 'more', 'than', 'twice', 'its', 'current', 'rate', 'indeed', 'a', 'few', 'economists', 'maintain', 'that', 'global', 'forces', 'will', 'continue', 'to', 'govern', 'hong', 'kong', "'s", 'economic', '<unk>', 'once', 'external', 'conditions', 'such', 'as', 'u.s.', 'demand', 'swing', 'in', 'the', 'territory', "'s", 'favor', 'they', 'argue', 'local', 'businessmen', 'will', 'probably', 'overcome', 'their', 'N', 'worries', 'and', 'continue', 'doing', 'business', 'as', 'usual', 'but', 'economic', 'arguments', 'however', 'solid', 'wo', "n't", 'necessarily', '<unk>', 'hong', 'kong', "'s", 'N', 'million', 'people', 'many', 'are', 'refugees', 'having', 'fled', 'china', "'s", '<unk>', 'cycles', 'of', 'political', 'repression', 'and', 'poverty', 'since', 'the', 'communist', 'party', 'took', 'power', 'in', 'N', 'as', 'a', 'result', 'many', 'of', 'those', 'now', 'planning', 'to', 'leave', 'hong', 'kong', 'ca', "n't", 'easily', 'be', '<unk>', 'by', '<unk>', 'improvements', 'in', 'the', 'colony', "'s", 'political', 'and', 'economic', 'climate' \\\\
}}
Figure on the middle: \blue{\textit{
'it', 'said', 'the', 'man', 'whom', 'it', 'did', 'not', 'name', 'had', 'been', 'found', 'to', 'have', 'the', 'disease', 'after', 'hospital', 'tests', 'once', 'the', 'disease', 'was', 'confirmed', 'all', 'the', 'man', "'s", 'associates', 'and', 'family', 'were', 'tested', 'but', 'none', 'have', 'so', 'far', 'been', 'found', 'to', 'have', 'aids', 'the', 'newspaper', 'said', 'the', 'man', 'had', 'for', 'a', 'long', 'time', 'had', 'a', 'chaotic', 'sex', 'life', 'including', 'relations', 'with', 'foreign', 'men', 'the', 'newspaper', 'said', 'the', 'polish', 'government', 'increased', 'home', 'electricity', 'charges', 'by', 'N', 'N', 'and', 'doubled', 'gas', 'prices', 'the', 'official', 'news', 'agency', '<unk>', 'said', 'the', 'increases', 'were', 'intended', 'to', 'bring', '<unk>', 'low', 'energy', 'charges', 'into', 'line', 'with', 'production', 'costs', 'and', 'compensate', 'for', 'a', 'rise', 'in', 'coal', 'prices', 'in', '<unk>', 'news', 'south', 'korea', 'in', 'establishing', 'diplomatic', 'ties', 'with', 'poland', 'yesterday', 'announced', '\$', 'N', 'million', 'in', 'loans', 'to', 'the', 'financially', 'strapped', 'warsaw', 'government', 'in', 'a', 'victory', 'for', 'environmentalists', 'hungary', "'s", 'parliament', 'terminated', 'a', 'multibillion-dollar', 'river', '<unk>', 'dam', 'being', 'built', 'by', '<unk>', 'firms', 'the', '<unk>', 'dam', 'was', 'designed', 'to', 'be', '<unk>', 'with', 'another', 'dam', 'now', 'nearly', 'complete', 'N', 'miles', '<unk>', 'in', 'czechoslovakia', 'in', 'ending', 'hungary', "'s", 'part', 'of', 'the', 'project', 'parliament', 'authorized', 'prime', 'minister', '<unk>', '<unk>', 'to', 'modify', 'a', 'N', 'agreement', 'with', 'czechoslovakia', 'which', 'still', 'wants', 'the', 'dam', 'to', 'be', 'built', 'mr.', '<unk>', 'said', 'in', 'parliament', 'that', 'czechoslovakia', 'and', 'hungary', 'would', 'suffer', 'environmental', 'damage', 'if', 'the', '<unk>', '<unk>', 'were', 'built', 'as', 'planned' \\\\
}}
Figure on the right: \blue{\textit{
'in', 'hartford', 'conn.', 'the', 'charter', 'oak', 'bridge', 'will', 'soon', 'be', 'replaced', 'the', '<unk>', '<unk>', 'from', 'its', '<unk>', '<unk>', 'to', 'a', 'park', '<unk>', 'are', 'possible', 'citizens', 'in', 'peninsula', 'ohio', 'upset', 'over', 'changes', 'to', 'a', 'bridge', 'negotiated', 'a', 'deal', 'the', 'bottom', 'half', 'of', 'the', '<unk>', 'will', 'be', 'type', 'f', 'while', 'the', 'top', 'half', 'will', 'have', 'the', 'old', 'bridge', "'s", '<unk>', 'pattern', 'similarly', 'highway', 'engineers', 'agreed', 'to', 'keep', 'the', 'old', '<unk>', 'on', 'the', 'key', 'bridge', 'in', 'washington', 'd.c.', 'as', 'long', 'as', 'they', 'could', 'install', 'a', 'crash', 'barrier', 'between', 'the', 'sidewalk', 'and', 'the', 'road', '<unk>', '<unk>', 'drink', 'carrier', 'competes', 'with', '<unk>', '<unk>', '<unk>', 'just', 'got', 'easier', 'or', 'so', 'claims', '<unk>', 'corp.', 'the', 'maker', 'of', 'the', '<unk>', 'the', 'chicago', 'company', "'s", 'beverage', 'carrier', 'meant', 'to', 'replace', '<unk>', '<unk>', 'at', '<unk>', 'stands', 'and', 'fast-food', 'outlets', 'resembles', 'the', 'plastic', '<unk>', 'used', 'on', '<unk>', 'of', 'beer', 'only', 'the', '<unk>', 'hang', 'from', 'a', '<unk>', 'of', '<unk>', 'the', 'new', 'carrier', 'can', '<unk>', 'as', 'many', 'as', 'four', '<unk>', 'at', 'once', 'inventor', '<unk>', 'marvin', 'says', 'his', 'design', 'virtually', '<unk>', '<unk>' \\\\
}}

\subsection{More generated text from the model:}\label{sec:appendix_text}
We illustrate below some generated text resulting from training 
\acrlong{TopicRNN} on the PTB dataset. Here we used $50$ neurons and $100$ topics: \\\\
Text1: \blue{\textit{but the refcorp bond fund might have been unk and unk of the point 
rate eos house in national unk wall restraint in the property pension fund 
sold willing to zenith was guaranteed by \$ N million at short-term 
rates maturities around unk products eos deposit posted yields slightly \\\\
}}
Text2: \blue{\textit{it had happened by the treasury 's clinical fund month 
were under national disappear institutions but secretary nicholas 
instruments succeed eos and investors age far compound average 
new york stock exchange bonds typically sold \$ N shares in 
the N but paying yields further an average rate of long-term funds\\\\
}}
We illustrate below some generated text resulting from training 
\acrlong{TopicRNN} on the IMDB dataset. The settings are 
the same as for the sentiment analysis experiment: \\\\
\blue{\textit{the film 's greatest unk unk and it will likely very nice 
movies to go to unk why various david proves eos the story were 
always well scary friend high can be a very strange unk unk is in love 
with it lacks even perfect for unk for some of the worst movies come 
on a unk gave a rock unk eos whatever let 's possible eos that kyle 
can 't different reasons about the unk and was not what you 're not a 
fan of unk unk us rock which unk still in unk 's music unk one as\\\\
}}

 \subsection{Topics from IMDB:}\label{sec:appendix_imdb_topics}
 Below we show some topics resulting from the sentiment analysis on the IMDB dataset.
 The total number of topics is $200$. Note here all the topics turn around movies which 
 is expected since all reviews are about movies.
\begin{table}[!hbpt]
\caption{\label{tab:topics_imdb} Some Topics from the TopicRNN Model on the IMDB Data.}
\begin{center}
\resizebox{\columnwidth}{!}{
\begin{tabular}{c|c|c|c|c|c|c}
pitt  & tarantino & producing & ken & hudson & campbell & campbell \\
cameron & dramas & popcorn & opera & dragged & africa & spots \\
vicious & cards & practice & carrey & robinson & circumstances & dollar \\
francisco & unbearable & ninja & kong & flight & burton & cage \\
los & catches & cruise & hills & awake & kubrick & freeman \\
revolution & nonsensical & intimate & useless & rolled & friday & murphy \\
refuses & cringe & costs & lie & easier & expression & $2002$ \\
cheese & lynch & alongside & repeated & kurosawa & struck & scorcese \\
\end{tabular}}
\end{center}
\end{table}

\end{document}